\documentclass[10pt,twocolumn]{article}
\usepackage{booktabs}
\usepackage{cvpr}
\usepackage{times}
\usepackage{epsfig}
\usepackage{graphicx}
\usepackage{amsmath}
\usepackage{amssymb}
\usepackage{subfigure}
\usepackage{graphicx}

\usepackage{flushend}

\cvprfinalcopy

\begin{document}

\title{Estimating Uncertainty in Landslide Segmentation Models}

\author{Savinay Nagendra $^{1}$  \quad Chaopeng Shen$^{2}$ \quad Daniel Kifer$^{1}$ \vspace{0.3em} \\
{\normalsize $^1$Department of Computer Science} \quad
{\normalsize $^2$Department of Civil and Environmental Engineering} \quad \\
{\normalsize The Pennsylvania State University}\\
{\normalsize University Park}\\
{\tt\small\centering $^1$sxn265@psu.edu \quad $^2$cshen@engr.psu.edu \quad $^1$dkifer@cse.psu.edu \vspace{0.3em}}}

\maketitle
\begin{abstract}
Landslides are a recurring, widespread hazard. Preparation and mitigation efforts can be aided by a high-quality, large-scale dataset that covers global at-risk areas. Such a dataset currently does not exist and is impossible to construct manually. Recent automated efforts focus on deep learning models for landslide segmentation (pixel labeling) from satellite imagery. However, it is also important to characterize the uncertainty or confidence levels of such segmentations. Accurate and robust uncertainty estimates can enable low-cost (in terms of manual labor) oversight of auto-generated landslide databases to resolve errors, identify hard negative examples, and increase the size of labeled training data. In this paper, we evaluate several methods for assessing pixel-level uncertainty of the segmentation. Three methods that do not require architectural changes were compared, including Pre-Threshold activations, Monte-Carlo Dropout and Test-Time Augmentation -- a method that measures the robustness of predictions in the face of data augmentation. Experimentally, the quality of the latter method was consistently higher than the others across a variety of models and metrics in our dataset.


\end{abstract}

\section{Introduction}\label{sec:intro}
\begin{figure*}[!t]
    \centering
    \includegraphics[width=\textwidth]{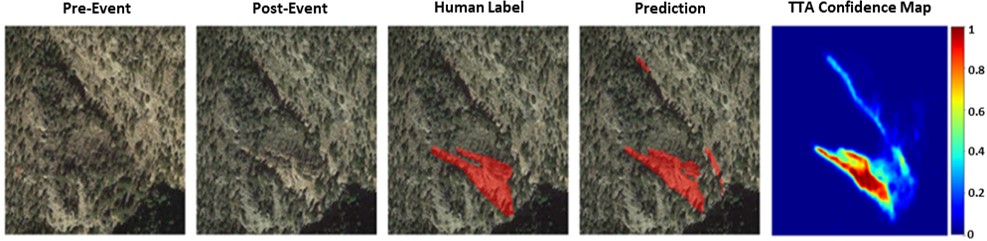}
    \vspace{-5pt}
    \caption{(Left to right) Example of bi-temporal image from our dataset, human label, prediction of our best model (Section \ref{sec:model}) and the corresponding TTA Confidence Map (Section \ref{sec:uncertainty})}
    \label{fig:intro_img}
\end{figure*}
Landslides are one of the most widespread hazards, with 300M people exposed and 66M people in high-risk areas \cite{article}. Globally, landslides causes thousands of deaths each year ($\approx$4,164 in 2017 alone \cite{petley_2018}) and the displacement of communities and destruction of roads and habitable lands. Moreover, climate change is projected to induce more frequent extreme rainfalls and wildfires, which are expected to result in more landslides and landslide-related casualties \cite{stocker_alexander_allen_2013,fischer_knutti_2015}.

Long-term efforts to mitigate the impacts of landslides require evaluating landslide susceptibility. Planning and prediction capabilities can be greatly assisted by a large-scale database of satellite images of landslide events, with accurate location information. In the US, the United States Geological Survey (USGS) maintains a database of landslide reports with approximate locations and times (but no images). However, even this kind of dataset, the most extensive of its kind, is incomplete and imprecise in location because it only gives an approximate location. While each entry can be verified manually -- either in-person or by examining satellite imagery, it presents both an opportunity and a challenge for scalability and coverage: ``our current ability to understand landslide hazards at the national scale is limited, in part because spatial data on landslide occurrence across the U.S. varies greatly in quality, accessibility, and extent'' \cite{https://doi.org/10.5066/p9e2a37p}. 

Given the increasing availability of high-resolution satellite images, a path is now open to collect information about space-visible landslide events using satellite images. Several recent efforts \cite{ghorbanzadeh2019evaluation,lei_zhang_lv_li_liu_nandi_2019} have considered the use of deep learning for landslide segmentation (also known as landslide mapping) -- identifying the pixels in a satellite image that correspond to landslides. However, due to the difficulty of labeling the training data, each study focused on on a single small region and had limited testing data (for instance, one study only had 3 testing images \cite{lei_zhang_lv_li_liu_nandi_2019}).

There are two types of tasks. The \emph{multi-image} task takes a current image (post-event image) and an earlier image (pre-event image) with the goal of identifying the landslide in the post-event image (hence it deals with image pairs). The  \emph{single-image} task simply takes one image and identifies the landslides (if any exist) in the image. It is useful when high quality \emph{recent} pre-event images are not available (for example, due to cloud cover).

Manually labeling training data does not scale as it is 
a challenging problem even for humans (see Figure \ref{fig:intro_img} for an example). Landslides are usually not the salient features in an image and there are often ``distractor'' objects (like roads) with similar color and texture. We found that manual annotation required training the labelers, who had to carefully examine each satellite image and compare it to the pre-event image. This process takes a human annotator several minutes per image.

Automated and semi-automated approaches for constructing landslide databases appear more promising as one could take an initial dataset, train landslide segmentation models and use them to segment additional satellite images. These new images could be fed back into the training data, models could be retrained, etc.  However, if auto-segmented images are fed back into this training loop, one risks propagating errors -- models trained on the new data will learn to make the same mistakes as the previous models while possibly adding new mistakes of their own. Thus human curators must play a role in detecting and correcting errors.

The key to making such an approach work is to accurately assess the pixel-level confidence of the segmentation models -- \textbf{for each pixel, how confident is the model that this is a landslide pixel?} Accurate confidence estimates can allow curators to prioritize the images they need to check, and contours of the uncertainty maps can be used to suggest alternative segmentations for an image (providing regions that can be added to or removed from an existing polygon).

%

Typical segmentation models provide a score for every pixel. These scores are then thresholded -- values above the threshold are labeled as landslide and values below are labeled as background. In principle, these pre-threshold values can be used as confidence measures but in practice there is much room for improvement. In this paper, we compare three methods (which do not change the underlying model architecture) for measuring per-pixel uncertainty: 
\begin{itemize}
    \item Pre-Threshold values -- these serve as a useful baseline and are computationally the fastest.
    \item Monte-Carlo Dropout \cite{gal2016dropout} -- this method is applicable to all models that are trained with dropout \cite{JMLR:v15:srivastava14a} (since dropout is a useful regularization technique, we consider it a training technique rather than an architectural change).
    \item Test-Time Augmentation (TTA) \cite{wang2019aleatoric} -- for a test image, we prepare multiple augmented images (rotated, flipped, etc.). A pixel's confidence score is the average of the pre-threshold values it receives from each of the augmented images. This technique makes the uncertainty scores robust in the sense that they are now (approximately) invariant to augmentations. To the best of our knowledge, this technique has not been used before in the remote sensing domain. 
\end{itemize}

We evaluate the confidence scores in three different ways. (1) We create calibration plots which are used to evaluate whether the confidence scores appear to act like probabilities  -- e.g., of all of the pixels having confidence 0.8 of being landslide, are 80\% (or above) of them actually landslide pixels? The next two measures evaluate how well the landslide confidence scores order the pixels. (2) Area Under the Curve (AUC) can be interpreted as the probability that a randomly chosen landslide pixel has a larger landslide-confidence score than a randomly chosen non-landslide pixel. (3) We also consider an adaptation of AUC to typical segmentation metrics like Intersection Over Union (IOU) -- if we take a confidence map for an image and sweep out all possible values of the threshold, what is the largest IOU that can be achieved for the image (we call this \textbf{image-specific thresholding}).

We find that across a variety of segmentation models, Test-Time Augmentation consistently outperforms the other methods in all three metrics.


This paper is organized as follows. We describe our landslide dataset in Section \ref{sec:approach}. Our evaluation uses 4 standard deep learning architectures, so in Section \ref{sec:model}, we explain what they are and present a brief comparison of their accuracies. In Section \ref{sec:uncertainty}, we describe the three techniques (for generating confidence maps) that are being evaluated for each of the deep learning architectures. In Section \ref{sec:evalmetrics}, we describe our evaluation criteria. Results appear in Section \ref{sec:results}. We present conclusions in Section \ref{sec:conclusion}.

\section{Related Work}\label{sec:related work}

While there has been prior work on detecting landslides from satellite images \cite{chen2018automated,ullo2019landslide,ghorbanzadeh2019evaluation,lei_zhang_lv_li_liu_nandi_2019}, none considered the use of uncertainty measurement. Due to difficulty in data collection, they use small datasets. Our dataset of 461 image pairs (one image before a landslide event and another after) is by far larger than prior work.

Previous work on estimating model uncertainty has primarily focused on two methods \cite{segalman2018epistemic}: a) aleatoric estimation which involves using the same model for multiple measurements while introducing randomness in other factors like input noise, and b) epistemic uncertainty which involves measuring the confidence of the model itself. While the latter can be achieved using Monte-Carlo (MC) Dropout \cite{gal2016dropout} essentially giving us a different model for each run using dropout during inference, the former has only been used for medical studies \cite{wang2019aleatoric}. Several studies \cite{NIPS2017_7141,liu2019accurate} showed that epistemic uncertainty can largely be explained away by a larger dataset, and it is more effective (but harder) to model aleatoric uncertainty. Often, uncertainty estimation using Bayesian approaches \cite{maddox2019simple,neal2012bayesian,graves2011practical,mackay1992practical} is considered the gold standard but, modern deep learning methods contain millions of parameters which makes the posterior highly non-convex in nature. Additionally, Bayesian analysis often requires significant changes to the training procedure, which makes uncertainty estimation computationally expensive. As an alternative to Bayesian approaches, MC Dropout is a simple and scalable technique motivated by approximate Bayesian inference. However, studies have shown that MC Dropout \cite{lakshminarayanan2017simple,maddox2019simple} tends to be over-confident and is exemplified in our empirical results. Deep Ensembles \cite{nagendra2017comparison, nagendra2022constructing, funk2018learning,liu2021new,pei2021utilizing,nagendra2020cloud,nagendra2020efficient,nagendra2022threshnet,nagendra2024patchrefinenet,zhu2022rapid} are also explored as a method of uncertainty estimation \cite{lakshminarayanan2017simple}, but is similar to using test-time dropout and requires training of multiple models.

The  method of aleatoric estimation (confidence maps) using test-time image augmentations \cite{wang2019aleatoric} tests the model for scale, rotation and blur invariance at the input level. Since aleatoric uncertainty cannot be reduced for a model \cite{liu2019accurate}, we essentially need to find the best model with minimal aleatoric uncertainty (Table \ref{tab:res3}) whose epistemic uncertainty can later be reduced by gathering a larger dataset \cite{NIPS2017_7141}. Our results show that test-time augmentations can provide as good or better results than MC Dropout and serves as a simple alternative to Bayesian/non-Bayesian/hybrid methods for segmentation tasks in remote sensing. For proving the validity of our uncertainty measure, multiple experiments such as calibration plots \cite{zadrozny2001obtaining}, AUC-ROC and image-specific thresholding are used as discussed in Section \ref{sec:uncertainty} and compared with MC Dropout and Pre-Threshold maps.

\section{Landslide Dataset}\label{sec:approach}


The datasets used in prior work have several limitations: (1) They
are predominantly small scale and contain limited landslide events (2) They use low resolution images where the landslide features are non-discriminative (3) They are spatially homogeneous, i.e., images are collected from a small geographical location. These factors will result in biased datasets, affecting the model's ability to generalize and could lead to over-fitting. To alleviate these problems and due to lack of a suitable alternative, we collect landslide images using the landslide inventory from United States Geological Survey (USGS).  \cite{https://doi.org/10.5066/p9e2a37p}. 

\begin{figure}[!t]
    \centering
    \includegraphics[width=7.68cm,height=5cm]{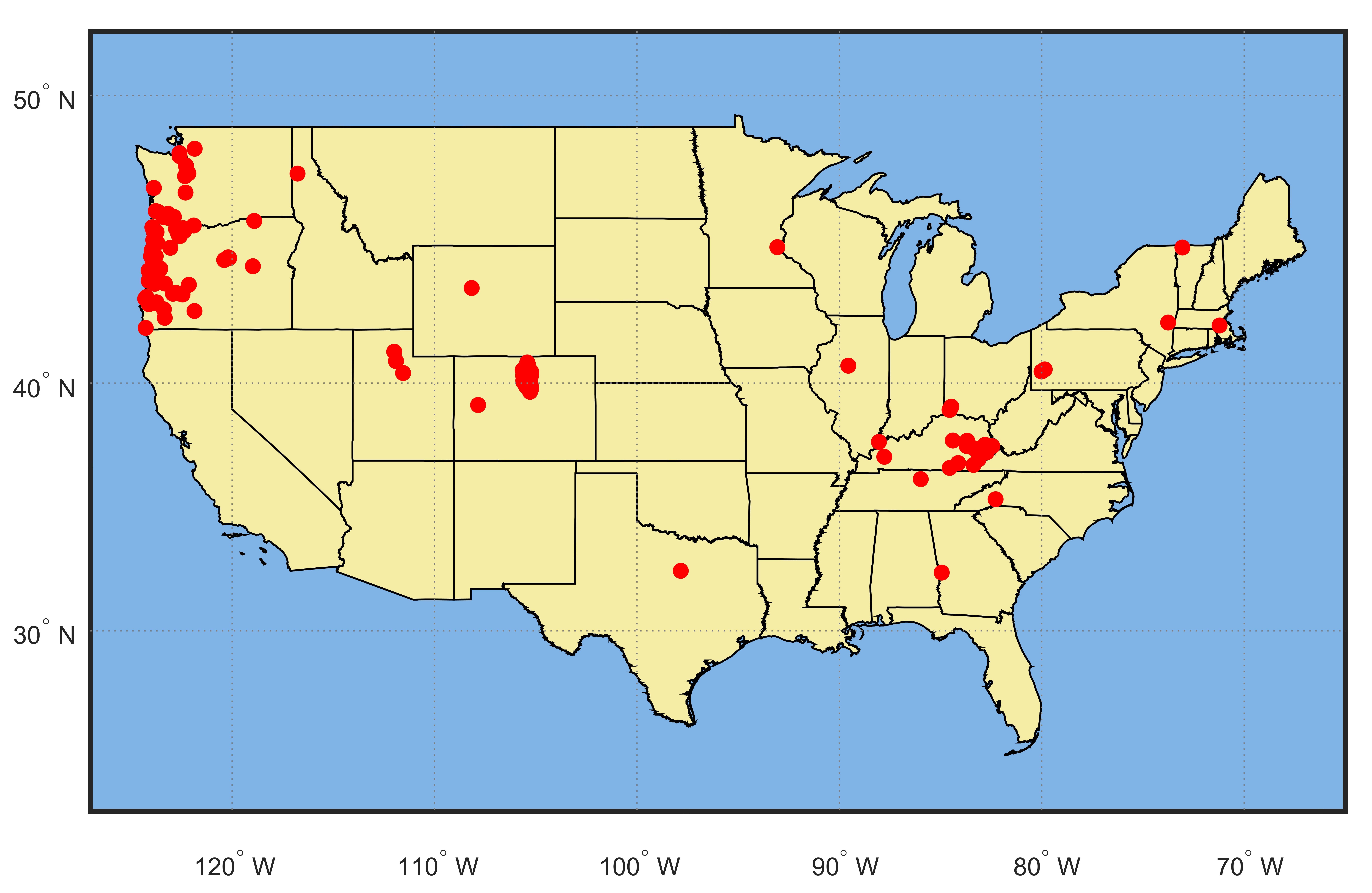}
    \caption{Distribution of verified space-visible landslides in the dataset.}
    \label{fig:Distribution of varified landslide events_v5}
\end{figure}
%
\begin{figure*}[!ht]
    \centering
     \includegraphics[width=\textwidth ]{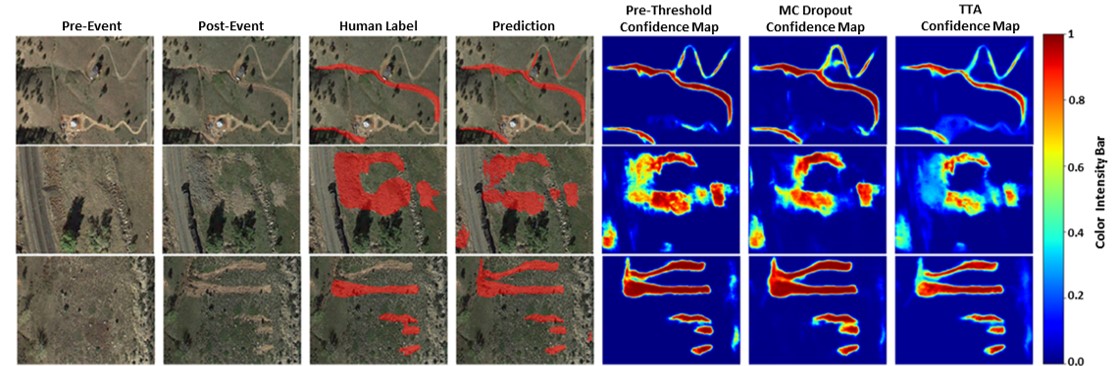}
    \caption{Qualitative results of U-Net (Multi) (best performing model) on example images from our dataset along with the three confidence maps evaluated.}
    \label{fig:eccv}
\end{figure*}

This inventory provides  approximate locations and times of landslide events. We examined these locations using Google Earth to find space-visible landslides -- the only ones we could verify remotely. 
We downloaded images for 1,120 verified landslide events (multiple events can occur in the same image). Majority of them were disregarded because of non-visibility using satellite images. This resulted in 461 pairs of bi-temporal images: one post-event image with a paired pre-event image. All the images were geo-referenced. 

Figure \ref{fig:Distribution of varified landslide events_v5} shows the distribution of verified space-visible landslides (red dots). It should be noted that our goal for this initial dataset is to only include rainfall-induced landslides. Landslides that occurred in California are excluded because many of the events were triggered by earthquakes. 
Figure \ref{fig:eccv} presents examples of bi-temporal images from our dataset along with segmentations from a UNet model and confidence maps for this model that were generated by 3 different methods. The images show diversity of terrains on which the landslides occurred. The confidence map generated by TTA (Section \ref{sec:uncertainty}), which performed best empirically, shows low confidence for the false positive areas in the top right and bottom left in the first and second rows, respectively. Note that in the third row,  the human labeler missed the top landslide patch, which is correctly predicted by the model. This shows the promise of using automated technology for the creation of landslide databases. 

For uniformity in our comparative study, we reshape all the images to 512$\times$512$\times$3. We perform data augmentation (flips, rotations, translations, random crops, shearing and gaussian blurs) for three purposes: First, data augmentations are known to improve performance of deep learning models by desensitizing the networks to unwanted perturbations. Second, data augmentations help eliminate the inherent unintentional bias in the dataset. For instance, the satellite images extracted from Google Earth are predominantly centered around landslides. Thus, we use random horizontal and vertical translations to remove this bias. Third, due to low saliency of landslide areas in satellite images, the number of foreground (landslide) pixels are far fewer than the number of background pixels, causing a pixel-wise class imbalance. Data augmentation reduces this imbalance.   

\section{Comparative Study}\label{sec:model}
Our confidence map comparison uses 4 deep learning models: FCN-8 \cite{long2015fully}, SegNet \cite{badrinarayanan2017segnet}, DeepLab \cite{chen2017deeplab}, and U-Net \cite{ronneberger2015u}. All models were trained from scratch, without pre-trained weights. For fair comparison between the two tasks (single-image and multi-image task), a model is trained for the multi-image task using the same capacity (number of parameters) as for the single-image task (same with hyper-parameter values and number of epochs). In this section, we establish the performance of these models (e.g., prediction accuracy) and in Section \ref{sec:uncertainty}, we use them to evaluate the different confidence map techniques. The quantitative comparison of the models is shown in Table \ref{tab:res2} and Table \ref{tab:res1}. A qualitative comparison is shown in Figure \ref{fig:results}.

\begin{table*}[!t]\centering
\resizebox{\linewidth}{!}{
\begin{tabular}{rrrrrrr} \hline
          
\textbf{Method} & \multicolumn{1}{c}{\textbf{mean IoU}(\%)} & \multicolumn{1}{c}{\textbf{mean Precision}(\%)} & \textbf{mean Recall}(\%) & \textbf{mean F1}(\%) & \textbf{mean Accuracy}(\%) & \textbf{\#Parameters} \\ \hline

FCN-8 \cite{long2015fully} & 57.6 & 79.3 & 69.1 & 72.2& 95.6& 134,326,918 \\
SegNet \cite{badrinarayanan2017segnet}& 62.4 & 85.2&70.8 &75.9 & 96.4& 31,820,801 \\
DeepLab\cite{chen2017deeplab} & 64.9 & 82.3 & 76.5 & 78.2& 96.6 & 33,077,031 \\
U-Net \cite{ronneberger2015u}& \textbf{68.8} & \textbf{86.4}& \textbf{78.1} & \textbf{81.9}& \textbf{97.2} & 8,642,273\\\hline

\end{tabular}
}
\vspace{1pt} 
\caption{Quantitative results on \textit{multi-image} task.}
\label{tab:res2} \vspace{-10pt} 
\end{table*}

\begin{table*}[!t]\centering
\resizebox{\linewidth}{!}{
\begin{tabular}{rrrrrrr} \hline
          
\textbf{Method} & \multicolumn{1}{c}{\textbf{mean IoU}(\%)} & \multicolumn{1}{c}{\textbf{mean Precision}(\%)} & \textbf{mean Recall}(\%) & \textbf{mean F1}(\%) & \textbf{mean Accuracy}(\%) & \textbf{\#Parameters} \\ \hline

FCN-8 \cite{long2015fully} & 44.5 & 74.7&53.8 &60.2 &94.2 & 134,326,918 \\
SegNet \cite{badrinarayanan2017segnet}& 57.9 & 80.6 &68.2 &72.2 &95.6 & 31,820,801 \\
DeepLab\cite{chen2017deeplab} &59.9 &\textbf{82.4} & 68.7& 73.4&96.3 & 33,077,031 \\
U-Net \cite{ronneberger2015u}&\textbf{65.8} & 80.1& \textbf{79.55} &\textbf{79.82}&\textbf{96.35} &8,642,273 \\\hline

\end{tabular}
}
\vspace{1pt} 
\caption{Quantitative results on \textit{single-image} task.}
\label{tab:res1} \vspace{-10pt} 
\end{table*}
\subsection{Quantitative Comparison}
We calculated five metrics for the quantitative evaluation of the networks: Intersection over Union, Precision, Recall, F1 Score, Pixel-level Accuracy. They are standard metrics for evaluating semantic segmentation models. They are derived from TP (True Positives), 
FP (False Positives), TN (True Negatives) and FN (False Negatives) at the pixel-level.


Results are portrayed in Table \ref{tab:res2} and Table \ref{tab:res1}. The best \textit{mIoU} achieved is 65.8\% on single-image task and 68.8\% on multi-image task. The results show the efficiency of multi-image models to capture temporal information despite having the same capacity (number of parameters) as single-image models. This is because the multi-image setting enables the models to find the difference between the post-event and pre-event images. This difference helps models to efficiently segment landslide regions in the post-event images. It is interesting to see that while U-Net \cite{ronneberger2015u} has a high recall in single-image task, it does not have a very high precision score. This indicates that the U-Net model mis-classifies neighboring regions (roads, boulders, water streams) as landslide regions in the single-image task, but with the multi-image task it learns to eliminate these false positives and thus the overall precision increases. U-Net \cite{ronneberger2015u}, SegNet \cite{badrinarayanan2017segnet} and FCN-8 \cite{long2015fully} show great improvements in precision from single-image to multi-image setting whereas DeepLab \cite{chen2017deeplab} does not. Conclusively, U-Net \cite{ronneberger2015u} is the best model for both the tasks followed closely by DeepLab \cite{chen2017deeplab} (in single-image) and SegNet \cite{badrinarayanan2017segnet} (in multi-image). 

\begin{figure}
    \centering
    \includegraphics[width=\linewidth]{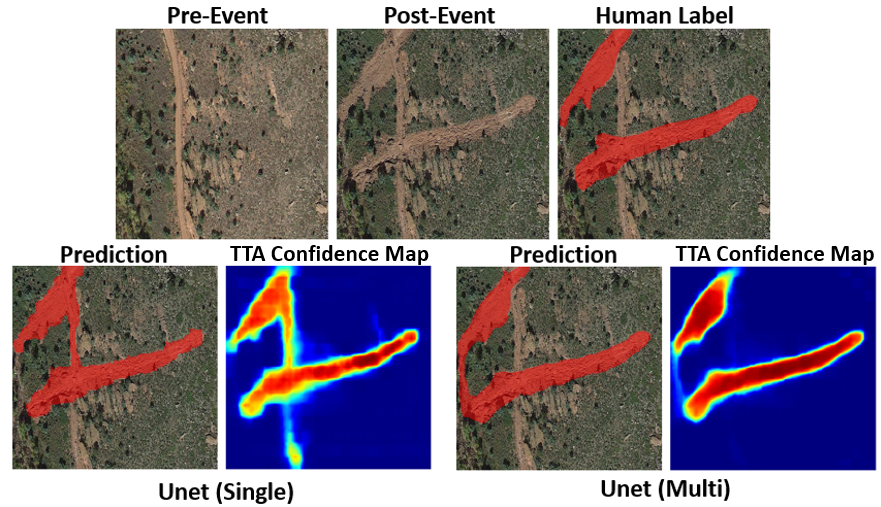}
    \caption{Qualitative results of U-Net (single-image and multi-image task)}
    \label{fig:qual}
\end{figure}

\subsection{Qualitative Comparison}
\begin{figure*}[!ht]
    \centering
    \includegraphics[width=\textwidth,height=15cm]{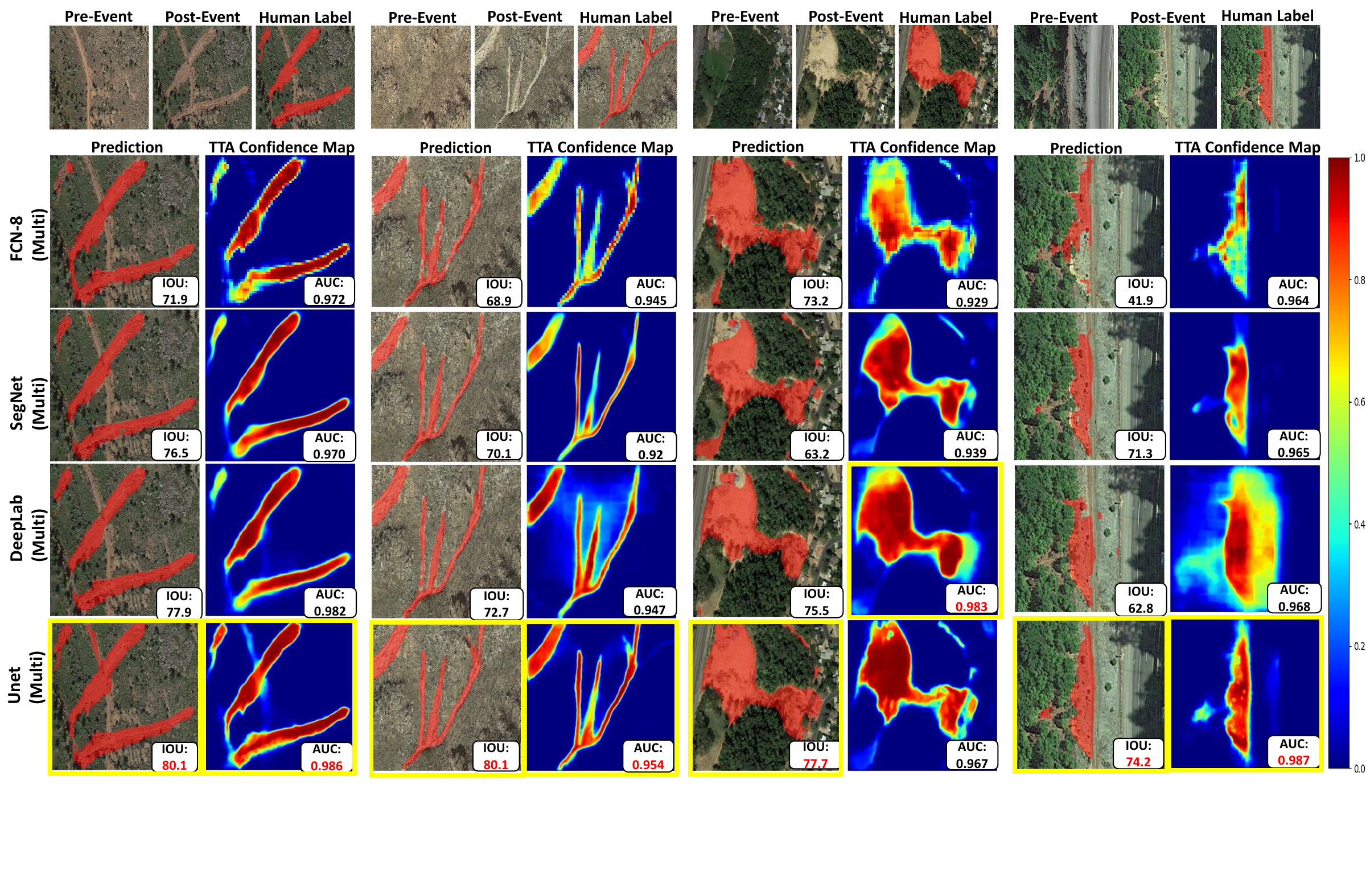}
    \vspace{-60 pt}
    \caption{Qualitative comparison of the models}
    \label{fig:results}
\end{figure*}

To qualitatively assess the predictions, we visualize the pixel-wise outputs of all the models in Figure \ref{fig:results}. An important thing to note is that several objects (roads, boulders, water streams -- false positive cases) in the gathered images are similar to landslides due to comparatively low resolution, thereby reducing the saliency of landslide pixels and making them more difficult to segment in the post-event image.

A close observation of Figure \ref{fig:qual} shows that the multi-image model for U-Net performs better than the single-image model by ignoring the neighbouring regions of similar pixel intensity. Usually, multi-image U-Net is more accurate than the other multi-image methods for detecting landslide pixels and ignoring neighboring pixels. 
%

In the following sections, for each of these models, we compare the different uncertainty quantification techniques to see which ones make it easier to detect model mistakes.
\section{Representing Confidence}\label{sec:uncertainty}
In this section we describe three methods that estimate model uncertainty: Pre-threshold values, Monte Carlo dropout \cite{gal2016dropout}, and Test-Time augmentations (TTA) \cite{wang2019aleatoric}. These methods provide a confidence score for each pixel, as shown in Figure \ref{fig:eccv}. 

\subsection{Pre-Threshold Maps} 

To obtain Pre-Threshold maps, we take the values that are output for each pixel (we call this the pixel intensity). Instead of thresholding these values (which is done to create a segmentation), we can simply use them to represent the confidence of the model. This can be justified, since the pixel intensities are numbers between 0 and 1 and the objective function during model training is binary cross-entropy with the human labels. We note that the pixel intensity values are expected to be crowded at the extremes. Theoretically, this is not a good way of representing confidence in predictions, but it serves as a baseline for comparison  with the other two methods.

\subsection{Monte-Carlo Dropout Maps}
Monte-Carlo Dropout \cite{gal2016dropout} is a method for estimating epistemic uncertainty in the models. High uncertainty scores using this method either implies over-fitting or limited training examples. It does so by inferencing the model with a different dropout mask every time. This is represented as $f_{nn}^{d_1}(x)$...$f_{nn}^{d_T}(x)$, where \textit{T} = 286, $x$ is the input image, and $f_{nn}^{d_i}$ represents the model output (\textbf{after thresholding}) with dropout mask $d_i$. 

For each test image we find {$f_{nn}^{d_T}(x)$} and have 286 maps corresponding to different dropout masks. Then, we compute the average of these masks to find the ensemble prediction ($\mu$), as shown in Equation \ref{eq:mean}. 
\begin{equation}
    \mu = \frac{\sum_{i=0}^{T} f_{nn}^{d_T}(x)}{T}
    \label{eq:mean}
\end{equation}
As this is a post-threshold average, it is equivalent to counting the fraction of times (across the different dropouts) that the pixel was thresholded to 1. Hence it is an estimate of the probability that the pixel should be thresholded to 1 (i.e. classified as landslide). This process is repeated for each of the 97 test images.

\subsection{Test-Time Augmentation Maps}
In this method we perform \textit{N = 286} different test-time augmentations on each test image. 
They are generated as follows. First we have a set of 6 geometric augmentations: horizontal flip, vertical flip, 2 diagonal flips and rotations of $\pm 90$ degrees. We also have a set of 40 vision-related augmentations: 20 Gaussian blurs, 10 linear contrasts and 10 brightness values. For each image, we consider all pairs of one geometric and one visual distortion (giving 6x40=240), a single geometric distortion by itself (6) and a single visual distortion by itself (40), giving a total of 286.


Once we have our augmented test set ready, we evaluate our segmentation model on these images. To match the spatial coherence of the original predictions, reverse augmentations are performed on the model predictions for the geometric augmentations (i.e., if an image was flipped horizontally, the predicted mask is flipped back to the original orientation). The mean over all the 286 predictions will generate a confidence map for a single test image (Equation \ref{eq:conf}, with $T=286$). This process is repeated for each of the 97 test images.
\begin{equation}
    \text{Confidence score (for pixel p)} (C_p) = \frac{\sum_{t=1}^{T} p_t}{T}
    \label{eq:conf}
\end{equation}

\section{Evaluation Criteria}\label{sec:evalmetrics}

In this section we explain how we evaluate the quality of the confidence maps discussed in section \ref{sec:uncertainty}.

\begin{table*}[!t] \centering
\resizebox{\linewidth}{!}{
  \begin{tabular}{ccccccc}
 \toprule
 &\multicolumn{3}{c}{{\textbf{IoU\textsubscript{a}}} \textbf{(using Image-Specific Thresholding)}}
 &
 \multicolumn{3}{c}{\textbf{AUC}} \\\cmidrule(r){2-4}\cmidrule(l){5-7}
 \textbf{Method} &\textbf{Pre-Threshold Map}& \textbf{Monte-Carlo Dropout Map}&\textbf{TTA Map}   &\textbf{Pre-Threshold Map}& \textbf{Monte-Carlo Dropout Map}&\textbf{TTA Map}      \\
 \hline
  FCN-8 \cite{long2015fully} & 0.628 &  0.611  & \textbf{0.677}&  0.903 & 0.898   &\textbf{0.938}\\
 SegNet \cite{badrinarayanan2017segnet}& 0.676 & 0.692 & \textbf{0.703} & 0.910 & 0.933 & \textbf{0.945}\\
  DeepLab \cite{chen2017deeplab}&0.684 & 0.684 & \textbf{0.708}&0.934 & 0.945 &\textbf{0.948}\\
  U-Net \cite{ronneberger2015u}& 0.723& 0.728& \textbf{0.73}& 0.951 & 0.948 &\textbf{0.96}\\
 \bottomrule
 \end{tabular}
}
\vspace{1pt} 
\caption{Results on \textit{multi-image} task.}
\label{tab:res3} \vspace{-10pt} 
\end{table*}

\begin{table*}[ht!] \centering
\resizebox{\linewidth}{!}{
 \begin{tabular}{ccccccc}
\toprule
 &\multicolumn{3}{c}{{\textbf{IoU\textsubscript{a}}} \textbf{(using Image-Specific Thresholding)}}
&
\multicolumn{3}{c}{\textbf{AUC}} \\\cmidrule(r){2-4}\cmidrule(l){5-7}
\textbf{Method} &\textbf{Pre-Threshold Map}& \textbf{Monte-Carlo Dropout Map}&\textbf{TTA Map}   &\textbf{Pre-Threshold Map}& \textbf{Monte-Carlo Dropout Map}&\textbf{TTA Map}      \\
\hline
 FCN-8 \cite{long2015fully} & 0.514  & 0.558   &\textbf{0.580}&0.812 & 0.833 &\textbf{0.835}\\
 SegNet \cite{badrinarayanan2017segnet}&0.623 & 0.641&\textbf{0.647} &0.852 & 0.866 &\textbf{0.872}\\
 DeepLab \cite{chen2017deeplab}&0.643 & 0.626&\textbf{0.661}&0.926 & 0.915&\textbf{0.935}\\
 U-Net \cite{ronneberger2015u}& 0.633&0.677 & \textbf{0.691}&0.922&0.930&\textbf{0.943} \\
\bottomrule
\end{tabular}
}
\vspace{1pt} 
\caption{Results on \textit{single-image} task.}
\label{tab:res4} \vspace{-10pt} 
\end{table*}

\subsection{Calibration Plots}
For comparing the uncertainty measures, we compute the calibration plots (Figure \ref{fig:cal}) for each method. Calibration plots are an intuitive way of indicating if the uncertainty measure is over-confident about the predicted pixels. To compute the calibration plot of an uncertainty method, we take its confidence map and bin the pixels based on their confidence of being a landslide (for example, we create a bin of the pixels whose confidence was in the range 0.8-0.9). For each bin, we then compute the fraction of actual landslide pixels in the bin (based on ground-truth labels). We bin the entire range [0,1] with intervals of 0.1. For a \textit{well-calibrated map}, this fraction of landslide pixels should lie within the bin boundaries, i.e. for the bin [0.9-1.0],  the percentage of actual  landslide pixels should also be in between [90\% - 100\%]. An ideal uncertainty measure would thus be the line \textit{y = x} (gold-standard) as shown in Figure \ref{fig:cal}. An under-confident model would start below this line, cross it at 0.5, and then go above the line (a sigmoid shape). An over-confident map would start above the line, then cross it and finish below the line. All else being equal, under-confidence is preferable to over-confidence (as over-confidence is not helpful for detecting errors).

We plot calibration maps as line charts for each multi-image model in order to compare the quality  of the three uncertainty measures. The method that follows the characteristic of a \textit{``well-calibrated map"} is regarded as the better method.

\subsection{Area Under Curve (AUC)}
For the three uncertainty methods, we find the Area underneath the ROC curve (receiver operating characteristic curve). This provides an aggregate measure of the uncertainty method's performance across all possible confidence thresholds [0-1]. It is equivalent to the probability that a randomly selected landslide pixel has higher confidence than a randomly selected background pixel. Higher the AUC is, better the uncertainty method is.

\subsection{Image-Specific Thresholding}

In this method, we take the confidence map of a test image $i$ and, by sweeping out different thresholds ($t^{(i)}=0.1, 0.2, ,\dots, 0.9$), we find the one that results in the best IoU (and then we average this number across the $N=97$ test images, see Equation \ref{eq:per}).
\begin{equation}
    IoU_a = \frac{\sum_{i=1}^{N}max(IoU_{t^{(i)}_1}, IoU_{t^{(i)}_2} ...IoU_{t^{(i)}_m})}{N}\label{eq:per}
\end{equation}
 We note that sweeping out the threshold is analogous to the procedure of computing an ROC curve. One way to view this metric is to consider a data labeler who is trying to correct a modeling mistake by choosing among the contours of the uncertainty map.  Another way to view it is that this metric, like AUC, estimates the tendency for a landslide pixel to have a higher landslide confidence score than the background pixels. Intuitively, if the pixel intensity distribution of the foreground and background are not well separated, the maximum achievable IoU score for any threshold would be low. The higher the value of maximum IoU (i.e., $IoU_a$), the better is the uncertainty method.
\section{Results} \label{sec:results}

First, we perform image-specific thresholding in Table \ref{tab:res3} and Table \ref{tab:res4} and infer that the $IoU_a$ score over the test set is highest in the case of TTA maps (0.73), and the worst for Pre-Threshold maps (0.723). Higher value in the case of TTA indicates that the landslide pixels are well separated from the background pixels in the confidence map, thus making TTA a better method than the other two. 
\begin{figure*}[!t]
    \centering     
    \includegraphics[width=\textwidth, height=105mm]{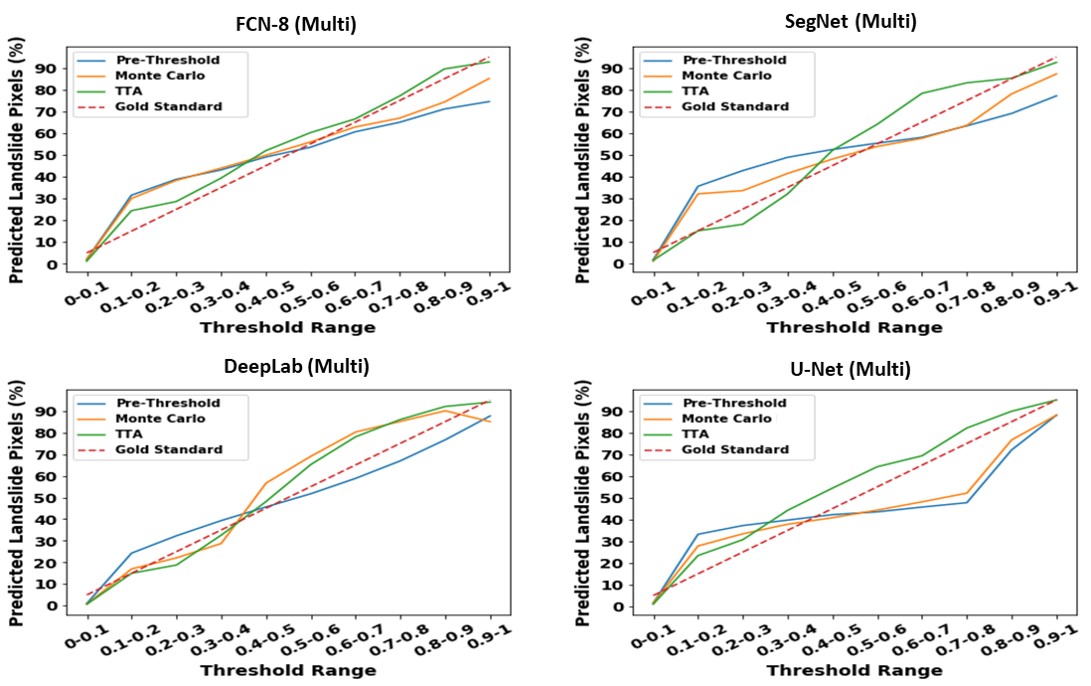}
    \caption{Calibration plots for the three uncertainty methods. TTA is closest to the gold standard for U-Net and FCN-8. For SegNet and DeepLab, TTA shows under-confident results. Monte-Carlo and Pre-Threshold usually shows over-confident results}
    \label{fig:cal}
\end{figure*}

To further check our understanding, we evaluate the Area under Curve (AUC) score for each method (Table \ref{tab:res3} and Table \ref{tab:res4}) and find that it is highest in the case of TTA maps (0.96), and the worst for Monte-Carlo maps (0.948). Higher value of AUC for TTA shows its efficiency in providing more confidence to a randomly selected landslide pixel compared to a randomly selected background pixel.

Finally we evaluate the calibration plots for the three methods. Figure \ref{fig:cal}, shows the line-charts of the three uncertainty measure. A ``well-calibrated" map should be close to the gold-standard. Whereas an over-confident map typically is above the gold-standard line in [0-0.5] range and below the gold standard in [0.5-1] range,
while the under-confident being the exact opposite. From Figure \ref{fig:cal}, we can observe that for FCN-8 \cite{long2015fully} and U-Net \cite{ronneberger2015u}, TTA exhibits the behaviour of a ``well-calibrated" map, whereas for DeepLab \cite{chen2017deeplab} and SegNet \cite{badrinarayanan2017segnet}
it is an under-confident map. Pre-Threshold maps and Monte-Carlo Dropout maps are usually over-confident for all the models. TTA appears to be under-confident for certain models, but all things being equal, it may be easier to detect possible errors made by a landslide annotation tool using an under-confident classifier.

\begin{figure}[!t]
    \centering     
    \includegraphics[width=\linewidth, height=46mm]{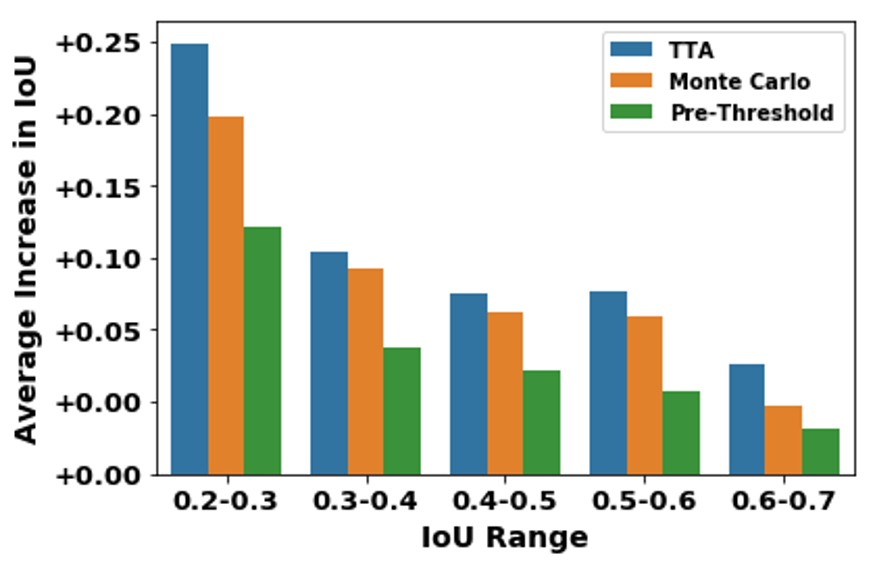}
    \caption{Average Increase in IoU for given IoU ranges for the three uncertainty methods.}
    \label{fig:iou_percent}
\end{figure}

Although, according to calibration plots TTA is not unambiguously the best measure, we can infer that the problem of landslide detection can be formulated as a multitask setting where a sub-network can be trained for finding the optimal threshold of each TTA map for gain in quantitative metrics. We simulate such an approach in Figure \ref{fig:iou_percent} in which, we calculate the average gain in the IoU scores using the best threshold for each image. The average gain is calculated for IOU ranges of length 0.1 (some ranges at the extremes contain almost no images and end up being very noisy; thus they are omitted due to their small sample size). Highest gains in each range is observed using TTA, thus indicating the separability of classes in the confidence map. 

The above experiments confirm our understanding of TTA being a \textit{``better"} method of uncertainty measurement for our dataset. Specifically, Table \ref{tab:res3} and Table \ref{tab:res4} demonstrate the efficacy of TTA across all the deep learning models (FCN-8 \cite{long2015fully}, SegNet \cite{badrinarayanan2017segnet}, DeepLab \cite{chen2017deeplab}, and U-Net \cite{ronneberger2015u}) for both the tasks (single-image and multi-image). Furthermore, we can infer from Table \ref{tab:res2}, Table \ref{tab:res1}, Table \ref{tab:res3} and Table \ref{tab:res4} that landslide detection can be approached as a \textit{multi-image} segmentation task for efficient results. U-Net for multi-image turns out to be the best overall model and is most certain about its predictions across all models based on $IoU_a$ and $AUC$ scores.

\section{Conclusion}\label{sec:conclusion}

In this paper, we introduce the first comparative study of uncertainty of segmentation models on a newly gathered landslide dataset. We demonstrate the effectiveness of the trained models for segmenting the landslide regions using uncertainty methods. Our experiments show that U-Net is the best performing model for both tasks (single-image and multi-image) with a high mIoU, yet it may not be confident about its decisions (similarly for the other models). While doing so, we discover that confidence maps created using test-time augmentations is effective in estimating the uncertainty of our models which can be crucial for creating accurate landslide annotation tools and has the potential to be extended to other tasks involving semantic classification of UAV/aerial and satellite images and videos. The ability of TTA to produce confidence maps that are approximately invariant to rotation, scale, blur and other variations at the input level makes it an effective method to estimate  aleatoric uncertainty.


{\small
\bibliographystyle{ieee_fullname}
\bibliography{egbib}

\begin{thebibliography}{10}\itemsep=-1pt

\bibitem{badrinarayanan2017segnet}
Vijay Badrinarayanan, Alex Kendall, and Roberto Cipolla.
\newblock Segnet: A deep convolutional encoder-decoder architecture for image
  segmentation.
\newblock {\em IEEE transactions on pattern analysis and machine intelligence},
  39(12):2481--2495, 2017.

\bibitem{chen2017deeplab}
Liang-Chieh Chen, George Papandreou, Iasonas Kokkinos, Kevin Murphy, and Alan~L
  Yuille.
\newblock Deeplab: Semantic image segmentation with deep convolutional nets,
  atrous convolution, and fully connected crfs.
\newblock {\em IEEE transactions on pattern analysis and machine intelligence},
  40(4):834--848, 2017.

\bibitem{chen2018automated}
Zhong Chen, Yifei Zhang, Chao Ouyang, Feng Zhang, and Jie Ma.
\newblock Automated landslides detection for mountain cities using
  multi-temporal remote sensing imagery.
\newblock {\em Sensors}, 18(3):821, 2018.

\bibitem{article}
M. Dilley, R.S. Chen, U. Deichmann, A. Lerner-Lam, M. Arnold, J. Agwe, P. Buys,
  O. Kjekstad, Bradfield Lyon, and Greg Yetman.
\newblock Natural disaster hotspots: A global risk analysis.
\newblock {\em World Bank Disaster Risk Management Series}, 5:1--132, 01 2005.

\bibitem{fischer_knutti_2015}
E.~M. Fischer and R. Knutti.
\newblock Anthropogenic contribution to global occurrenceof heavy-precipitation
  andhigh-temperature extremes.
\newblock {\em Nature Climate Change}, 5(6):560–564, 2015.

\bibitem{funk2018learning}
Christopher Funk, Savinay Nagendra, Jesse Scott, Bharadwaj Ravichandran, John~H
  Challis, Robert~T Collins, and Yanxi Liu.
\newblock Learning dynamics from kinematics: Estimating 2d foot pressure maps
  from video frames.
\newblock {\em arXiv preprint arXiv:1811.12607}, 2018.

\bibitem{gal2016dropout}
Yarin Gal and Zoubin Ghahramani.
\newblock Dropout as a bayesian approximation: Representing model uncertainty
  in deep learning.
\newblock In {\em international conference on machine learning}, pages
  1050--1059, 2016.

\bibitem{ghorbanzadeh2019evaluation}
Omid Ghorbanzadeh, Thomas Blaschke, Khalil Gholamnia, Sansar~Raj Meena, Dirk
  Tiede, and Jagannath Aryal.
\newblock Evaluation of different machine learning methods and deep-learning
  convolutional neural networks for landslide detection.
\newblock {\em Remote Sensing}, 11(2):196, 2019.

\bibitem{graves2011practical}
Alex Graves.
\newblock Practical variational inference for neural networks.
\newblock In {\em Advances in neural information processing systems}, pages
  2348--2356, 2011.

\bibitem{https://doi.org/10.5066/p9e2a37p}
Eric~S Jones, Benjamin~B Mirus, Robert~G Schmitt, Rex~L Baum, Jonathan~W Godt,
  Dalia~B Kirschbaum, Thomas~A Stanley, and Kevin~E. MCCoy.
\newblock Summary metadata - landslide inventories across the united states,
  2019.

\bibitem{NIPS2017_7141}
Alex Kendall and Yarin Gal.
\newblock What uncertainties do we need in bayesian deep learning for computer
  vision?
\newblock In I. Guyon, U.~V. Luxburg, S. Bengio, H. Wallach, R. Fergus, S.
  Vishwanathan, and R. Garnett, editors, {\em Advances in Neural Information
  Processing Systems 30}, pages 5574--5584. Curran Associates, Inc., 2017.

\bibitem{lakshminarayanan2017simple}
Balaji Lakshminarayanan, Alexander Pritzel, and Charles Blundell.
\newblock Simple and scalable predictive uncertainty estimation using deep
  ensembles.
\newblock In {\em Advances in neural information processing systems}, pages
  6402--6413, 2017.

\bibitem{lei_zhang_lv_li_liu_nandi_2019}
Tao Lei, Yuxiao Zhang, Zhiyong Lv, Shuying Li, Shigang Liu, and Asoke~K. Nandi.
\newblock Landslide inventory mapping from bitemporal images using deep
  convolutional neural networks.
\newblock {\em IEEE Geoscience and Remote Sensing Letters}, 16(6):982–986,
  2019.

\bibitem{liu2019accurate}
Jeremiah Liu, John Paisley, Marianthi-Anna Kioumourtzoglou, and Brent Coull.
\newblock Accurate uncertainty estimation and decomposition in ensemble
  learning.
\newblock In {\em Advances in Neural Information Processing Systems}, pages
  8950--8961, 2019.

\bibitem{liu2021new}
Jiangtao Liu, Chaopeng Shen, Te Pei, Kathryn Lawson, Daniel Kifer, Savinay
  Nagendra, and Srikanth Banagere~Manjunatha.
\newblock A new rainfall-induced deep learning strategy for landslide
  susceptibility prediction.
\newblock In {\em AGU Fall Meeting Abstracts}, volume 2021, pages NH35E--0504,
  2021.

\bibitem{long2015fully}
Jonathan Long, Evan Shelhamer, and Trevor Darrell.
\newblock Fully convolutional networks for semantic segmentation.
\newblock In {\em Proceedings of the IEEE conference on computer vision and
  pattern recognition}, pages 3431--3440, 2015.

\bibitem{mackay1992practical}
David~JC MacKay.
\newblock A practical bayesian framework for backpropagation networks.
\newblock {\em Neural computation}, 4(3):448--472, 1992.

\bibitem{maddox2019simple}
Wesley~J Maddox, Pavel Izmailov, Timur Garipov, Dmitry~P Vetrov, and
  Andrew~Gordon Wilson.
\newblock A simple baseline for bayesian uncertainty in deep learning.
\newblock In {\em Advances in Neural Information Processing Systems}, pages
  13132--13143, 2019.

\bibitem{nagendra2020efficient}
Savinay Nagendra, S Banagere~Manjunatha, Chaopeng Shen, Daniel Kifer, and Te
  Pei.
\newblock An efficient deep learning mechanism for cross-region generalization
  of landslide events.
\newblock In {\em AGU Fall Meeting Abstracts}, volume 2020, pages NH030--0010,
  2020.

\bibitem{nagendra2024patchrefinenet}
Savinay Nagendra and Daniel Kifer.
\newblock Patchrefinenet: Improving binary segmentation by incorporating
  signals from optimal patch-wise binarization.
\newblock In {\em Proceedings of the IEEE/CVF Winter Conference on Applications
  of Computer Vision}, pages 1361--1372, 2024.

\bibitem{nagendra2022constructing}
Savinay Nagendra, Daniel Kifer, Benjamin Mirus, Te Pei, Kathryn Lawson,
  Srikanth~Banagere Manjunatha, Weixin Li, Hien Nguyen, Tong Qiu, Sarah Tran,
  et~al.
\newblock Constructing a large-scale landslide database across heterogeneous
  environments using task-specific model updates.
\newblock {\em IEEE Journal of Selected Topics in Applied Earth Observations
  and Remote Sensing}, 15:4349--4370, 2022.

\bibitem{nagendra2020cloud}
S Nagendra, T Pei, S Banagere~Manjunatha, G He, T Qiu, D Kifer, and C Shen.
\newblock Cloud-based interactive database management suite integrated with
  deep learning-based annotation tool for landslide mapping.
\newblock In {\em AGU Fall Meeting Abstracts}, volume 2020, pages NH030--0011,
  2020.

\bibitem{nagendra2017comparison}
Savinay Nagendra, Nikhil Podila, Rashmi Ugarakhod, and Koshy George.
\newblock Comparison of reinforcement learning algorithms applied to the
  cart-pole problem.
\newblock In {\em 2017 international conference on advances in computing,
  communications and informatics (ICACCI)}, pages 26--32. IEEE, 2017.

\bibitem{nagendra2022threshnet}
Savinay Nagendra, Chaopeng Shen, and Daniel Kifer.
\newblock Threshnet: Segmentation refinement inspired by region-specific
  thresholding.
\newblock {\em arXiv preprint arXiv:2211.06560}, 2, 2022.

\bibitem{neal2012bayesian}
Radford~M Neal.
\newblock {\em Bayesian learning for neural networks}, volume 118.
\newblock Springer Science \& Business Media, 2012.

\bibitem{pei2021utilizing}
Te Pei, Savinay Nagendra, Srikanth Banagere~Manjunatha, Guanlin He, Daniel
  Kifer, Tong Qiu, and Chaopeng Shen.
\newblock Utilizing an interactive ai-empowered web portal for landslide
  labeling for establishing a landslide database in washington state, usa.
\newblock In {\em EGU General Assembly Conference Abstracts}, pages
  EGU21--13974, 2021.

\bibitem{petley_2018}
Dave Petley.
\newblock An analysis of fatal landslides, and the resultant deaths, in 2017,
  Apr 2018.

\bibitem{ronneberger2015u}
Olaf Ronneberger, Philipp Fischer, and Thomas Brox.
\newblock U-net: Convolutional networks for biomedical image segmentation.
\newblock In {\em International Conference on Medical image computing and
  computer-assisted intervention}, pages 234--241. Springer, 2015.

\bibitem{segalman2018epistemic}
Daniel~J Segalman and Matthew~RW Brake.
\newblock Epistemic and aleatoric uncertainty in modeling.
\newblock In {\em The Mechanics of Jointed Structures}, pages 593--603.
  Springer, 2018.

\bibitem{JMLR:v15:srivastava14a}
Nitish Srivastava, Geoffrey Hinton, Alex Krizhevsky, Ilya Sutskever, and Ruslan
  Salakhutdinov.
\newblock Dropout: A simple way to prevent neural networks from overfitting.
\newblock {\em Journal of Machine Learning Research}, 15:1929--1958, 2014.

\bibitem{stocker_alexander_allen_2013}
Thomas Stocker, Lisa Alexander, and Myles Allen.
\newblock {\em Climate change 2013: the physical science basis: final draft
  underlying scientific-technical assessment: Working Group I contribution to
  the IPCC fifth assessment report}.
\newblock WMO, IPCC Secretariat, 2013.

\bibitem{ullo2019landslide}
Silvia~L Ullo, Maximillian~S Langenkamp, Tuomas~P Oikarinen, Maria~P DelRosso,
  Alessandro Sebastianelli, S Sica, et~al.
\newblock Landslide geohazard assessment with convolutional neural networks
  using sentinel-2 imagery data.
\newblock In {\em IGARSS 2019-2019 IEEE International Geoscience and Remote
  Sensing Symposium}, pages 9646--9649. IEEE, 2019.

\bibitem{wang2019aleatoric}
Guotai Wang, Wenqi Li, Michael Aertsen, Jan Deprest, S{\'e}bastien Ourselin,
  and Tom Vercauteren.
\newblock Aleatoric uncertainty estimation with test-time augmentation for
  medical image segmentation with convolutional neural networks.
\newblock {\em Neurocomputing}, 338:34--45, 2019.

\bibitem{zadrozny2001obtaining}
Bianca Zadrozny and Charles Elkan.
\newblock Obtaining calibrated probability estimates from decision trees and
  naive bayesian classifiers.
\newblock In {\em Icml}, volume~1, pages 609--616. Citeseer, 2001.

\bibitem{zhu2022rapid}
L Zhu, P Tilke, S Nagendra, M Etchebes, and M LeFranc.
\newblock A rapid and realistic 3d stratigraphic model generator conditioned on
  reference well log data.
\newblock In {\em Second EAGE Digitalization Conference and Exhibition}, volume
  2022, pages 1--5. European Association of Geoscientists \& Engineers, 2022.

\end{thebibliography}
}

\end{document}